\documentclass[
]{ceurart}

\usepackage{amsmath,amssymb}
\usepackage{booktabs}
\usepackage{array}
\graphicspath{{assets/}}

\newcommand{\cmark}{$\checkmark$}

\newcommand{\xmark}{--}

\begin{document}
\copyrightyear{2026}
\copyrightclause{Copyright for this paper by its authors.
  Use permitted under Creative Commons License Attribution 4.0
  International (CC BY 4.0).}

\conference{CLEF 2026 Working Notes, 21 -- 24 September 2026, Jena, Germany}

\title{Candidate-Constrained Retrieval-Augmented Generation for LongEval-RAG: System Design and Empirical Analysis}

\title[mode=sub]{LongEval-RAG at CLEF 2026}

\author[1]{Yingdong Yang}[%
orcid=0009-0007-5546-5293,
email=yingdongyang0305@outlook.com,
url=https://github.com/yyd859/,
]
\fnmark[1]
\cormark[1]

\author[1]{Haijian Wu}[%
orcid=0009-0007-0311-6235,
email=willwuhj@gmail.com,
url=https://wijowill.github.io/,
]
\fnmark[1]

\cortext[1]{Corresponding author.}
\fntext[1]{These authors contributed equally and share first authorship.}

\begin{abstract}
We present a candidate-constrained retrieval-augmented generation system for LongEval-RAG, where each query is associated with an organizer-provided candidate set and all retrieved evidence and final citations must remain within that set. The system combines deterministic provenance tracking with passage-based retrieval, deterministic query expansion, pseudo-relevance feedback (PRF), reciprocal rank fusion (RRF), lightweight evidence reranking, citation-aware evidence aggregation, and optional MiniLM sentence reranking. We evaluate ten pipeline variants using a primary organizer evaluation and a supplementary self-generated diagnostic protocol. The primary evaluation shows that the strongest balanced variant is \texttt{rule-minilm}: a rule-based chunking pipeline with query expansion, PRF, RRF, reranking, citation prior, and late MiniLM sentence selection. This variant obtains the highest BERTScore, retrieval precision, nugget coverage, and average grade among our submissions. The result suggests that the main gain does not come from more complex semantic or topic-shift chunking, but from pairing stable rule-based passages with sentence-level neural selection before generation. The supplementary LLM-judge evaluation remains useful for early diagnosis and additional analysis, but it emphasizes different systems than the primary gold-answer and nugget-based evaluation, highlighting the need for multi-metric RAG evaluation.
\end{abstract}

\begin{keywords}
  Retrieval-augmented generation \sep
  LongEval-RAG \sep
  Evidence ranking \sep
  Citation-grounded generation
\end{keywords}

\maketitle

\section{Introduction}
LongEval-RAG studies answer generation from an organizer-provided candidate set rather than unrestricted corpus search \cite{longeval2026task4,lncs_overview_longeval_2026,ceur_overview_longeval_2026}. This setting differs from open-domain RAG, where retrieval is typically performed against a large external corpus such as Wikipedia \cite{lewis2020rag,gao2023rag_survey}. Instead, each query is paired with an official list of candidate document IDs, and every retrieved passage, selected evidence item, and final citation must remain inside that list. In our system, the main retrieval-time selection unit is the passage: the pipeline ranks and selects passages derived from candidate documents, then selects sentence-level evidence from those passages for answer generation, while the final \texttt{references} list is derived from the source document IDs of the selected evidence.

This constraint makes the task especially suitable for analyzing the internal quality of a RAG system. Since broad corpus recall is removed from the problem definition, performance depends more directly on how well the system structures documents into passages, ranks evidence, allocates answer space, and preserves provenance from final claims back to source documents.

This paper makes three main contributions. First, we describe a complete candidate-constrained RAG pipeline for LongEval-RAG, covering data preparation, passage construction, retrieval, evidence selection, answer generation, and deterministic provenance tracking. Second, we report an ablation study of ten pipeline variants under the primary multi-metric organizer evaluation, while retaining a supplementary cross-provider LLM-judge protocol as a diagnostic view. Third, we identify the strongest primary-evaluation configuration: \texttt{rule-minilm}, which combines deterministic rule-based chunks with late MiniLM sentence reranking and achieves the best balance of semantic answer similarity, retrieval precision, nugget coverage, and average nugget grade among our submissions.

The primary evaluation points to a clear empirical message. Query expansion, PRF, and RRF remain important because they broaden lexical evidence recall inside the fixed candidate pool, following Query2doc-style expansion \cite{wang2023query2doc}, relevance-model ideas \cite{lavrenko2001relevance}, and reciprocal rank fusion \cite{cormack2009rrf}. The strongest primary-evaluation configuration, \texttt{rule-minilm}, builds on this retrieval foundation by keeping stable rule-based passages and applying MiniLM only at the late sentence-selection stage. This suggests that, in the current setup, neural matching is most useful for allocating the final evidence budget rather than for redefining upstream chunk boundaries with semantic or topic-shift heuristics. The supplementary LLM-judge evaluation assigns its highest mean score to \texttt{rrf\_no\_rerank}, but we treat that difference as a diagnostic evaluation disagreement rather than as the main empirical conclusion.

\section{Task Setting and Data Preparation}
\subsection{Query and Document Inputs}
The query input is stored as JSONL, with one query per line. Each record contains a query identifier, the query text, and the official candidate document IDs for that query. The implementation also supports several field aliases for robustness, for example mapping identifiers from fields such as \texttt{narrative\_id}, \texttt{query\_id}, \texttt{qid}, or \texttt{id}, and mapping query text from fields such as \texttt{narrative}, \texttt{query}, \texttt{text}, or \texttt{question}.

The current evaluation slice contains 47 queries. Each query is paired with exactly 10 official candidate documents, yielding 470 query-document instances in total. Thus, the candidate count is fixed across the slice rather than varying by query.

Document input is loaded from JSONL, JSON, CSV, TSV, or a directory containing those formats. Each document record should include a document ID, a title, and a usable text field, preferably \texttt{fullText}, along with an optional publication timestamp. The loader likewise supports multiple field aliases for document IDs.

\subsection{Candidate-Constrained Join Logic}
The entire pipeline is candidate-constrained by design. The loading process first reads all query records, extracts the candidate \texttt{doc\_ids} associated with each query, collects the union of required document IDs, and loads only those documents from the document collection. The pipeline then attaches documents back to each query in the order specified by its candidate list.

This design has an important methodological consequence: the system never retrieves outside the official candidate set. All later stages, including passage construction, retrieval, reranking, answer generation, and evaluation, operate only over the provided candidate pool.

\subsection{Text Selection and Provenance Tracking}
For text composition, the loader follows a simple priority rule:
\[
\texttt{fullText} \rightarrow \texttt{abstract} \rightarrow \texttt{title}.
\]
Whenever full text is available, it is used as the main evidence source. If full text is missing, the system falls back to the abstract, and then to the title if necessary. This fallback is an engineering decision for data availability rather than a preferred scientific choice.

Table~\ref{tab:text_availability} reports text availability for the current 47-query slice. Among the 470 query-document instances, 222 instances use full text, 248 use abstract-only fallback, and none require title-only fallback. No candidate documents are dropped for missing text in this slice.

\begin{table}[t]
\centering
\small
\caption{Text availability in the current evaluation slice. Percentages are computed over 470 query-document instances.}
\label{tab:text_availability}
\begin{tabular}{lrr}
\toprule
Text source used & Count & Percentage \\
\midrule
\texttt{fullText} & 222 & 47.2\% \\
Abstract only & 248 & 52.8\% \\
Title only & 0 & 0.0\% \\
\bottomrule
\end{tabular}
\end{table}

Timestamps are loaded and preserved as document metadata. In the retrieval configurations reported here, they can also serve as active retrieval-side signals when the query has temporal intent. Specifically, the evidence reranker includes a configurable temporal boost, evidence selection can encourage coverage of earlier and later documents for temporally phrased queries, and the extractive generator can construct temporal comparison sentences from timestamp-derived years. However, timestamps are not exposed directly in the final batch LLM prompts for the \texttt{openai\_batch\_gpt54\_top10} evaluation inputs.

A central implementation feature is deterministic provenance tracking. After retrieval and evidence selection, the system converts the source document IDs of selected passages into an ordered cited-document list, stored in the output field \texttt{references}. Final answer sentences do not cite raw document IDs directly. Instead, each sentence stores citation indices into this \texttt{references} list. The reconstruction path is therefore:
\[
\text{citation index} \rightarrow \texttt{references[index]} \rightarrow \text{document ID}.
\]
For batch generation, the pipeline adds an \texttt{evidence\_id} layer so that asynchronous model outputs can still be mapped back to sentence candidates, citation indices, and finally source documents. This engineering choice makes every cited sentence in the final JSONL output traceable to its supporting document, in the same general spirit as recent citation-grounded generation setups \cite{gao2023alce}.

\subsection{Terminology for Text and Evaluation Units}
We use five terms consistently throughout the paper. They are not interchangeable: documents define the candidate set, chunks describe segmentation, passages are retrieval units, \texttt{references} records cited document IDs, and nuggets are evaluation units.

\begin{table}[t]
\centering
\scriptsize
\setlength{\tabcolsep}{3pt}
\caption{Core terminology used for retrieval, provenance, and evaluation.}
\label{tab:terminology}
\resizebox{\textwidth}{!}{%
\begin{tabular}{p{0.16\textwidth}p{0.38\textwidth}p{0.38\textwidth}}
\toprule
Term & Definition & Role in the paper \\
\midrule
Document & One organizer-provided candidate document ID for query $q$, denoted $d\in\mathcal{D}_q$ & Defines the fixed candidate set; every retrieved passage, prompt item, and final citation must trace back to one of these documents. \\
Chunk & A contiguous group of sentences produced by rule-based, semantic, or topic-shift segmentation & Describes how a document is split before indexing; chunking strategy is one of the main ablation dimensions. \\
Passage & The indexed retrieval unit emitted from a chunk, with text, source document ID, title, timestamp metadata, and sentence offsets & Unit scored by BM25, RRF, reranking, and passage-level selection before prompt construction. \\
Reference & Ordered, deduplicated list of source document IDs from selected passages & Output field used by citation indices; distinct from the bibliography of cited papers. \\
Nugget & Minimal information unit expected in a correct answer & Evaluation unit used by organizer metrics such as nugget coverage and average grade. \\
\bottomrule
\end{tabular}%
}
\end{table}

\section{System Pipeline}
\subsection{Overview and Notation}
For a query $q$, let $\mathcal{D}_q=\{d_1,\ldots,d_{10}\}$ denote the organizer-provided candidate document set. The pipeline first normalizes these documents into passages and then performs retrieval only over
\[
\mathcal{P}_q=\bigcup_{d\in\mathcal{D}_q}\operatorname{chunk}(d),
\]
where every passage $p\in\mathcal{P}_q$ stores its text, source document $d(p)$, title, timestamp when available, and sentence offsets. The candidate constraint is therefore enforced before ranking: no passage outside $\mathcal{D}_q$ can enter retrieval, reranking, evidence selection, or the final prompt.

Table~\ref{tab:pipeline_parameters} summarizes the concrete constants used by the reported configurations. These are the experiment-level parameters exposed in the current pipeline. BM25 is used as a lexical ranker, but downstream RRF consumes ranks rather than raw BM25 magnitudes, so BM25 scores are not treated as cross-component weights in the ablation analysis.

\begin{table}[t]
\centering
\scriptsize
\setlength{\tabcolsep}{3pt}
\caption{Main implementation parameters for the reported system pipeline.}
\label{tab:pipeline_parameters}
\resizebox{\textwidth}{!}{%
\begin{tabular}{lll}
\toprule
Stage & Parameter & Value / behavior \\
\midrule
Rule chunking & \texttt{passage\_max\_words} & 140 words \\
 & \texttt{passage\_stride\_sentences} & 1 sentence overlap \\
Semantic chunking & Embedding model & \texttt{all-MiniLM-L6-v2} \\
 & \texttt{semantic\_merge\_threshold} & 0.72 cosine similarity \\
Topic-shift chunking & \texttt{topic\_shift\_boundary\_threshold} & 0.18 adjacent-sentence drift \\
Semantic/topic chunking & Minimum sentences before split & 1 sentence \\
 & Maximum sentence count & unset; only word budget applies \\
Lexical retrieval & BM25 baseline & original-query passage ranking \\
 & BM25 parameters & $k_1=1.5$, $b=0.75$; not tuned \\
Query expansion & Extracted query keywords & up to 12 unique non-stopword tokens \\
 & \texttt{query\_variant\_limit} & 5 variants including the original query \\
PRF & \texttt{prf\_top\_passages} & top 4 original-query BM25 passages \\
 & Feedback terms & top 6 raw-frequency terms \\
Fusion & RRF constant & $k=60$ \\
Reranking & \texttt{preselect\_k} & 24 passages \\
Evidence selection & \texttt{top\_k\_passages} & 20 passages \\
 & \texttt{per\_doc\_limit} & unset; no hard per-document cap \\
Citation prior & \texttt{citation\_graph\_boost} & 0.05 additive weight \\
Prompt construction & Candidate sentences per passage & up to first 5 sentences \\
 & \texttt{max\_selected\_answer\_candidates} & 10 sentence-level evidence items \\
Sentence reranking & Cross-encoder model & \texttt{cross-encoder/ms-marco-MiniLM-L-6-v2} \\
Answer generation & \texttt{max\_answer\_sentences} & 5 answer sentences \\
\bottomrule
\end{tabular}%
}
\end{table}

\subsection{Passage Construction}
The default chunker is a deterministic overlapping sentence-window procedure. Let a document be represented as a sentence sequence $(s_1,\ldots,s_n)$, and let $w(C)$ be the word count of a candidate chunk $C$. The rule-based chunker scans left to right, greedily appending the next sentence while
\[
w(C\cup\{s_i\})\leq 140.
\]
When the next sentence would exceed the limit, the current chunk is emitted as a passage. The next chunk is initialized with the final sentence of the previous chunk, giving a one-sentence overlap. The 140-word value is used as a shared upper bound across the reported rule-based, semantic, and topic-shift variants, but it is not tuned in this study. We therefore treat passage length as a hyperparameter for future work, where it should be evaluated together with the resulting chunk-length distribution and downstream retrieval quality.

The semantic chunking variant replaces fixed sentence windows with embedding-based merging. Each sentence $s_i$ is embedded with \texttt{all-MiniLM-L6-v2}. For a current chunk $C$, the chunk embedding is the mean of its sentence embeddings:
\[
\bar{e}(C)=\frac{1}{|C|}\sum_{s_i\in C} e(s_i).
\]
A candidate sentence $s_j$ is merged into $C$ only when both constraints hold:
\[
\cos\bigl(e(s_j),\bar{e}(C)\bigr)\geq 0.72
\quad\text{and}\quad
w(C\cup\{s_j\})\leq 140.
\]
Otherwise, $C$ is emitted and a new chunk begins at $s_j$. The minimum number of sentences before a similarity-based split is one, and no additional maximum sentence-count cap is active in the reported semantic runs.

The topic-shift variant uses the same sentence embeddings but detects local drift between adjacent sentences. For $i>1$, define
\[
\Delta_i=1-\cos\bigl(e(s_{i-1}),e(s_i)\bigr).
\]
A new chunk starts before $s_i$ when the current chunk already contains at least one sentence and $\Delta_i>0.18$, or when adding $s_i$ would exceed the 140-word limit. This gives a lightweight topic-boundary heuristic related to TextTiling-style segmentation \cite{hearst1997texttiling}, while preserving the same size budget as the rule-based chunker.

\subsection{Lexical Ranking and Query Variants}
The single-query baseline ranks passages with BM25 over the original query text \cite{robertson2009bm25}. For query variant $v$ and passage $p$, the BM25 scoring form is
\[
\operatorname{BM25}(v,p)=\sum_{t\in v}\operatorname{IDF}(t)
\frac{f(t,p)(k_1+1)}{f(t,p)+k_1\left(1-b+b\frac{|p|}{\operatorname{avgdl}}\right)},
\]
where $f(t,p)$ is the frequency of term $t$ in passage $p$, $|p|$ is passage length, $\operatorname{avgdl}$ is the average passage length in $\mathcal{P}_q$, and the reported runs use $k_1=1.5$ and $b=0.75$. These BM25 parameters are implementation constants rather than tuned hyperparameters. The reported ablations use BM25 to define ranked lists; once lists are produced, fusion depends on rank positions rather than raw BM25 scores.

Deterministic query expansion is fully local. It uses no LLM, external dictionary, or retrieval-derived text. The system tokenizes the original query narrative, lowercases tokens, removes stopwords and very short tokens, keeps only tokens with length at least three, deduplicates them by first occurrence, and preserves their original query order. It then keeps at most 12 such tokens as the query-keyword set. From this set, the system can construct a keyword-focused variant, a temporal-intent variant if the tokens overlap a small hand-written temporal-intent list, and a method-intent variant if the tokens overlap a small hand-written method-intent list. The total number of query variants is capped at five including the original query.

Pseudo-relevance feedback is a separate retrieval-derived expansion step. Let $F_q$ be the four highest-ranked passages under original-query BM25. A candidate feedback term must have length at least four, must not be a stopword, and must not already appear in the extracted query-keyword set. Its score is raw frequency in the feedback pool:
\[
\operatorname{PRF}(t,q)=\sum_{p\in F_q} f(t,p).
\]
The top six terms under this score are appended only to the query-keyword set to form one additional PRF query variant. They are not added back into the original, keyword, temporal, or method variants. This means PRF contributes one separate BM25 ranked list to RRF rather than changing all query variants. The separation keeps query-only expansion and retrieval-derived expansion distinguishable: original and deterministic variants reflect only the user query, while the PRF variant reflects terms found in the first-pass feedback passages.

\subsection{Rank Fusion}
The multi-query systems retrieve one lexical ranked list for each enabled variant: the original query, keyword query, optional temporal query, optional method query, and optional PRF query. Dense retrieval is supported by the implementation, but the reported Task 4 runs leave the dense model disabled, so the pre-reranking fusion is lexical-only. The ranked lists are merged with reciprocal rank fusion \cite{cormack2009rrf}:
\[
S_{\operatorname{RRF}}(p\mid q)=\sum_{r\in\mathcal{R}_q}\frac{1}{60+\operatorname{rank}_r(p)},
\]
where $\mathcal{R}_q$ is the set of enabled ranked lists for query $q$. If passage $p$ is absent from a list, that list contributes zero. The only RRF hyperparameter is the denominator offset $k=60$.

\subsection{Reranking and Citation Prior}
Reranked variants take the top 24 passages under $S_{\operatorname{RRF}}$ as a shortlist. The reranker is deliberately lightweight: it preserves the RRF score as the base ranking signal and adds a small document-level citation prior. For document $d$, let $c(d)$ be the number of edges in the provided citation network in which $d$ appears either as \texttt{citing\_doc\_id} or as \texttt{cited\_doc\_id}. The normalized citation prior is
\[
G(d)=\frac{c(d)}{\max_{d'}c(d')},
\]
so $G(d)\in[0,1]$. For a passage $p$ from document $d(p)$, the configured additive citation term is
\[
0.05\,G(d(p)).
\]
Thus the explicitly configured additive score used by the main reranked variants is
\[
S_{\operatorname{rerank}}(p\mid q)=S_{\operatorname{RRF}}(p\mid q)+0.05\,G(d(p))+T(q,p),
\]
where $T(q,p)$ denotes the temporal adjustment applied only for temporally phrased queries with usable document timestamps. The timestamp signal is used on the retrieval side to encourage earlier/later coverage, but timestamps are not printed in the final batch prompts.

\subsection{Evidence Selection and Prompt Construction}
After reranking, the system selects 20 passages per query. The main configurations do not impose a hard per-document quota: a document can contribute multiple passages if they remain highly ranked. The selector filters near-duplicate passages and, for temporal queries, may force coverage of the earliest and latest dated documents before filling the remaining passage slots from the ranked list. The source document IDs of the selected passages are then converted into the ordered \texttt{references} list, deduplicated by first occurrence.

The answer-generation prompt uses a smaller sentence-level evidence budget. The generator extracts up to the first five candidate sentences from each selected passage, filters and scores those sentences, and keeps at most 10 sentence-level evidence items. MiniLM sentence-reranking variants insert an additional cross-encoder scoring step over these sentence candidates; non-MiniLM variants retain the deterministic order induced by passage selection. Each sentence-level evidence item carries an \texttt{evidence\_id} and citation indices into the \texttt{references} list.

The \texttt{rule-minilm} variant is methodologically important because it separates passage construction from neural sentence selection. Passage boundaries remain deterministic and overlap-preserving, while MiniLM is used only after candidate evidence has been narrowed to sentence-level answer candidates. This design reduces the risk that embedding-based segmentation will discard useful local context, while still allowing a neural model to prioritize sentences that are semantically close to the query.

Unless otherwise noted, all answer generation runs use OpenAI \texttt{gpt-5.4-mini}. The model receives only the selected candidate-constrained evidence items and is limited to at most five answer sentences. After batch generation, returned \texttt{evidence\_id} values are mapped back to sentence candidates, citation indices, and source document IDs. This preserves the candidate constraint at the prompt level and makes every final citation traceable to an official candidate document.

\section{Experimental Variants}
Table~\ref{tab:methods} summarizes the ten experimental variants currently tracked in the repository. Together, they serve three purposes: retrieval-strength ablation, chunking-strategy comparison, and sentence-selection comparison.

\begin{table}[t]
\centering
\scriptsize
\setlength{\tabcolsep}{3pt}
\caption{Method matrix for the current LongEval Task 4 experiments. Abbreviations: Rule = rule chunking, Sem. = semantic chunking, Topic = topic-shift chunking, BM25 = single-query BM25, QE = query expansion, RR = evidence reranking, Cit. = citation prior, Sent. = MiniLM sentence reranking.}
\label{tab:methods}
\resizebox{\textwidth}{!}{%
\begin{tabular}{rlcccccccccc}
\toprule
No. & Approach & Rule & Sem. & Topic & BM25 & QE & PRF & RRF & RR & Cit. & Sent. \\
\midrule
1 & \texttt{concat-baseline} & \cmark & \xmark & \xmark & \xmark & \xmark & \xmark & \xmark & \xmark & \xmark & \xmark \\
2 & \texttt{single-query-bm25} & \cmark & \xmark & \xmark & \cmark & \xmark & \xmark & \xmark & \xmark & \xmark & \xmark \\
3 & \texttt{rrf-no-rerank} & \cmark & \xmark & \xmark & \xmark & \cmark & \cmark & \cmark & \xmark & \xmark & \xmark \\
4 & \texttt{caes-rag-rrf}$^{\ast}$ & \cmark & \xmark & \xmark & \xmark & \cmark & \cmark & \cmark & \cmark & \cmark & \xmark \\
5 & \texttt{default}$^{\ast}$ & \cmark & \xmark & \xmark & \xmark & \cmark & \cmark & \cmark & \cmark & \cmark & \xmark \\
6 & \texttt{rule-minilm} & \cmark & \xmark & \xmark & \xmark & \cmark & \cmark & \cmark & \cmark & \cmark & \cmark \\
7 & \texttt{semantic-current} & \xmark & \cmark & \xmark & \xmark & \cmark & \cmark & \cmark & \cmark & \cmark & \xmark \\
8 & \texttt{topic-shift-current} & \xmark & \xmark & \cmark & \xmark & \cmark & \cmark & \cmark & \cmark & \cmark & \xmark \\
9 & \texttt{semantic-minilm} & \xmark & \cmark & \xmark & \xmark & \cmark & \cmark & \cmark & \cmark & \cmark & \cmark \\
10 & \texttt{topic-shift-minilm} & \xmark & \xmark & \cmark & \xmark & \cmark & \cmark & \cmark & \cmark & \cmark & \cmark \\
\midrule
\multicolumn{12}{@{}l}{\footnotesize $^{\ast}$ \texttt{caes-rag-rrf} and \texttt{default} use the same effective setting and serve as a control pair.} \\
\bottomrule
\end{tabular}%
}
\end{table}

Table~\ref{tab:methods} should be read together with the following role summary. \texttt{concat\_baseline} is the lower bound, \texttt{single\_query\_bm25} is the lexical baseline, and \texttt{rrf\_no\_rerank} isolates the gain from multi-query retrieval. \texttt{caes\_rag\_rrf} and \texttt{default} form a control pair because they use the same effective setting: rule-based chunks, query expansion, PRF, RRF, evidence reranking, and citation prior, without MiniLM sentence reranking. The semantic and topic-shift variants are chunking ablations, while the MiniLM variants test late sentence-level reranking after passage retrieval.

The terminology is important because MiniLM appears in two different places. In the \texttt{semantic} and \texttt{topic-shift} chunking families, \texttt{all-MiniLM-L6-v2} embeddings define or detect chunk boundaries. In contrast, the \texttt{minilm} suffix denotes a later cross-encoder sentence reranker applied to already selected sentence candidates; \texttt{current} denotes the corresponding configuration without that sentence reranker. Thus, \texttt{rule-minilm} keeps rule-based chunks and adds neural scoring only at sentence allocation time.

Interpreting the matrix at a high level, the progression
\texttt{concat\_baseline} $\rightarrow$ \texttt{single\_query\_bm25} $\rightarrow$ \texttt{rrf\_no\_rerank} $\rightarrow$ \texttt{caes\_rag\_rrf}
serves as a staged comparison from no retrieval, to single-query lexical retrieval, to multi-query lexical fusion, and then to the full reranked variant. The semantic and topic-shift families test whether more sophisticated segmentation yields better retrieval passages, while the MiniLM variants test whether a compact Transformer model can serve as an additional reranker after retrieval \cite{wang2020minilm,nogueira2019passage_rerank}.

\section{Evaluation Protocols}
We use two evaluation protocols with different roles. The \textbf{primary evaluation} is the official organizer evaluation and provides the main empirical evidence in the paper. The \textbf{supplementary evaluation} is our self-generated cross-provider diagnostic protocol. It was developed much earlier than the official release, so it allowed us to compare candidate models, estimate rough system quality, and check usability and stability during system development.

\subsection{Primary Organizer Evaluation}
The primary evaluation is the organizer-released evaluation. The organizer report provides a broad evaluation package rather than a single score: it includes 33 available measures and recommends a compact subset that captures complementary aspects of retrieval-augmented generation. We report the measures that are most useful for interpreting our systems: ROUGE-L, BERTScore, retrieval precision, nugget coverage, average grade, and TFC1.

These metrics evaluate different failure modes. ROUGE-L is reported as \texttt{rougeL\_f1}: it computes longest-common-subsequence overlap between the generated answer and the organizer-provided gold answer, then reports the harmonic mean of ROUGE-L precision and recall \cite{lin2004rouge}. BERTScore is reported as \texttt{bertscore\_f1}: generated-answer tokens and gold-answer tokens are matched with contextual embeddings, and the final value is the harmonic mean of BERTScore precision and recall \cite{zhang2020bertscore}. Retrieval precision is the fraction of cited documents that are judged relevant to the query, so a run that cites only relevant documents receives precision 1.0 and a run that cites only non-relevant documents receives 0.0. Nugget coverage is the recall of minimal information nuggets covered by the response, and average grade is the mean rubric grade assigned to those nuggets; the organizer implementation uses TREC-AutoJudge-style nugget evaluation, where nuggets are derived from preference comparisons and then judged for response coverage \cite{dietz2026nugget_banks,farzi2026autojudge}. TFC1 is an axiomatic RAG measure that gives preference to responses with more occurrences of query terms, following the RAG axiom framework implemented in TREC-AutoJudge \cite{merker2025rag_axioms,farzi2026autojudge}.

We use these metrics together because each one isolates a different part of the RAG pipeline. A system can have good ROUGE-L by matching surface wording without using the best evidence; it can have high retrieval precision while still omitting important answer nuggets; and it can improve query-term concentration without improving semantic completeness. Therefore, the primary evaluation is not treated as a simple leaderboard, but as a multi-view diagnostic of answer similarity, citation quality, nugget coverage, and query-focused evidence use.

This interpretation is especially important for LongEval-RAG because all runs operate inside the same fixed candidate-document set. Since broad corpus recall is not the main variable, differences between systems are more likely to reflect passage construction, evidence allocation, sentence selection, and citation behavior. For this reason, we treat the primary metrics as the main empirical basis for comparing variants, while using the supplementary protocol to explain disagreements and inspect provenance behavior.

\subsection{Supplementary Diagnostic Evaluation}
The supplementary evaluation is an internal diagnostic protocol that we built before the primary organizer results were available. Its first purpose was practical: it gave us an early way to compare variants, select promising configurations, and monitor whether the end-to-end system was usable and stable before the organizer-side gold-answer and nugget-based evaluation was released.

\begin{figure}[t]
\centering
\includegraphics[width=0.92\linewidth]{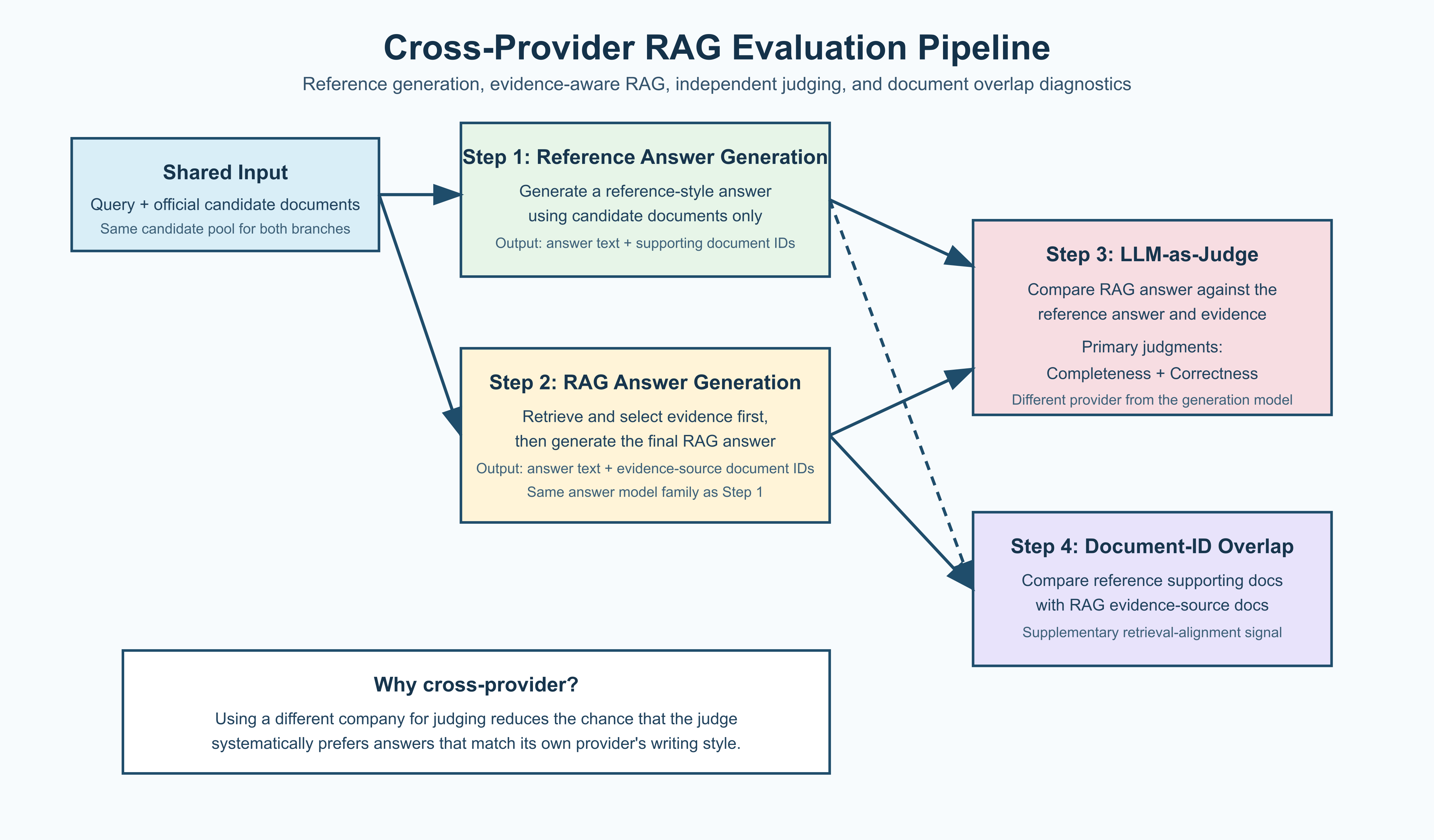}
\caption{Overall pipeline used in our supplementary LongEval Task 4 experiments, covering candidate-constrained retrieval, answer generation, and cross-provider evaluation.}
\label{fig:pipeline}
\end{figure}

Its second purpose is complementary measurement. As summarized in Figure~\ref{fig:pipeline}, the supplementary protocol generates a reference-style answer with document-ID attribution, generates a candidate-constrained RAG answer, scores the answer with a separate LLM judge (Claude Sonnet 4.5) and computes document-ID overlap diagnostics. These outputs expose dimensions that are useful for development and error analysis, including completeness, correctness, citation-set overlap, reference-subset matching, unsupported citations, and provenance consistency.

Table~\ref{tab:summary-by-model} reports six supplementary metrics, each averaged over the 47-query slice. \textbf{Corr.} is the Claude Sonnet 4.5 judge's correctness score: it measures whether the RAG answer is factually consistent with the reference answer and available evidence, without unsupported claims. \textbf{Comp.} is the same judge's completeness score: it measures how much of the reference answer's important information is covered. Both judge scores use the same bounded rubric, and \textbf{Avg.} is their arithmetic mean, $(\mathrm{Corr.}+\mathrm{Comp.})/2$. This follows the general LLM-as-judge setup used in recent generation-evaluation work, while retaining a cross-provider separation between answer generator and judge \cite{zheng2023llm_judge,liu2023geval}. Let $G$ be the reference-side cited document set and $R$ be the RAG answer's cited document set. \textbf{Recall} is $|G\cap R|/|G|$, \textbf{Precision} is $|G\cap R|/|R|$, and \textbf{Jaccard} is $|G\cap R|/|G\cup R|$, the standard Jaccard set-overlap coefficient \cite{jaccard1901distribution}. These citation-set metrics are diagnostic rather than final answer-quality measures: they indicate whether the system grounds its answer in documents similar to those used by the reference side.

Its third purpose is evaluation analysis. By comparing the supplementary results with the primary results, we can analyze both the systems and the evaluation protocols themselves. On the system side, the comparison shows which variants look strong under a fluency- and completeness-oriented local judge but weaker under gold-answer, retrieval-precision, and nugget-based measures. On the evaluation side, the disagreement helps reveal what each protocol rewards, where LLM-as-judge evaluation may overvalue plausible synthesis, and why RAG systems should be discussed with multiple complementary metrics rather than a single score.

\section{Primary Results and Analysis}
Table~\ref{tab:primary-evaluation} reports the primary organizer scores for the submitted runs. We keep this view distinct from the supplementary evaluation: \emph{primary evaluation} refers to organizer-released metrics, whereas \emph{supplementary evaluation} refers to our self-generated cross-provider judge and overlap diagnostics.

\begin{table}[t]
\centering
\scriptsize
\setlength{\tabcolsep}{3pt}
\caption{Primary organizer evaluation for Task 4 of LongEval. We report the F$_{1}$ score of ROUGE-L and BERTScore as primary measures. Additionally, we report retrieval precision, nugget coverage, average grade, and TFC1. A dash indicates that the corresponding value was not present in the local organizer-results export used for this table.}
\label{tab:primary-evaluation}
\resizebox{\textwidth}{!}{%
\begin{tabular}{@{}rlcccccc@{}}
\toprule
\bf No. & \bf Approach & \bf ROUGE-L & \bf BERTScore & \bf Precision & \bf Nugget Cov. & \bf Avg. Grade & \bf TFC1 \\
\midrule
1 & \texttt{rule-minilm} & 0.157 & \textbf{0.168} & \textbf{0.975} & \textbf{0.367} & \textbf{2.929} & 0.305 \\
2 & \texttt{topic-shift-minilm} & 0.157 & 0.163 & 0.940 & 0.288 & 2.671 & \textbf{0.373} \\
3 & \texttt{semantic-minilm} & 0.160 & 0.157 & 0.922 & 0.291 & 2.696 & 0.349 \\
4 & \texttt{topic-shift-current} & \textbf{0.167} & 0.156 & 0.929 & 0.292 & 2.694 & 0.215 \\
5 & \texttt{semantic-current} & 0.156 & 0.150 & 0.933 & 0.282 & 2.651 & 0.262 \\
6 & \texttt{single-query-bm25} & 0.156 & 0.152 & 0.940 & 0.308 & 2.744 & 0.120 \\
7 & \texttt{rrf-no-rerank} & 0.155 & 0.151 & 0.929 & 0.336 & 2.816 & 0.087 \\
8 & \texttt{caes-rag-rrf} & 0.155 & 0.144 & 0.942 & 0.338 & 2.830 & 0.059 \\
9 & \texttt{default} & 0.147 & 0.146 & 0.931 & 0.334 & 2.845 & 0.025 \\
10 & \texttt{official-naive-baseline} & 0.082 & -0.118 & 0.200 & 0.124 & 1.828 & -0.758 \\
\bottomrule
\end{tabular}%
}
\end{table}

For clarity, each Table~\ref{tab:primary-evaluation} metric is computed as follows. ROUGE-L is \texttt{rougeL\_f1}, the F$_1$ form of longest-common-subsequence overlap with the gold answer \cite{lin2004rouge}. BERTScore is \texttt{bertscore\_f1}, the F$_1$ form of contextual-embedding token matching with the gold answer \cite{zhang2020bertscore}. Precision is cited-document precision, nugget coverage is covered-nugget recall, average grade is the mean nugget rubric grade, and TFC1 is the query-term-frequency RAG axiom from the organizer's TREC-AutoJudge-style evaluation \cite{dietz2026nugget_banks,farzi2026autojudge,merker2025rag_axioms}.

All submitted retrieval-based systems substantially outperform the official naive baseline, especially on retrieval precision. The main analysis therefore focuses less on whether retrieval is useful at all and more on which evidence-allocation strategy works best inside the fixed candidate pool.

\subsection{Primary Evaluation Favors Rule-Based Chunking With Late MiniLM Reranking}
The strongest balanced primary-evaluation run is \texttt{rule-minilm}. It achieves the best BERTScore (0.168), retrieval precision (0.975), nugget coverage (0.367), and average grade (2.929) among our submitted systems. Its ROUGE-L score is not the highest, but its advantage across semantic similarity, cited-evidence precision, and nugget-oriented measures makes it the most convincing overall configuration under the primary protocol.

This result suggests that the best intervention in the current pipeline is not upstream embedding-based segmentation. Instead, the most effective configuration keeps deterministic rule-based chunks, uses query expansion, PRF, RRF, lightweight reranking, and a citation prior to form a strong candidate pool, and then applies MiniLM late to allocate the final sentence-level evidence budget.

\subsection{Different Primary Metrics Reward Different Behaviors}
The primary metrics do not identify a single winner on every axis. \texttt{topic-shift-current} obtains the highest ROUGE-L score (0.167), suggesting that topic-shift chunking can improve surface lexical overlap with the reference answer. \texttt{topic-shift-minilm} obtains the highest TFC1 score (0.373), suggesting stronger query-term or factual-constraint concentration in selected evidence. However, neither run matches \texttt{rule-minilm} on the combination of BERTScore, retrieval precision, nugget coverage, and average grade.

This metric spread is useful because it shows that end-to-end RAG quality should not be reduced to a single lexical-overlap number. Lexical similarity, semantic similarity, cited-evidence precision, nugget coverage, and query-term faithfulness each expose a different part of the pipeline.

\subsection{Current vs. MiniLM: Late Sentence Reranking Is Helpful But Not Uniform}
Within the semantic-chunking family, adding MiniLM sentence reranking improves ROUGE-L from 0.156 to 0.160, BERTScore from 0.150 to 0.157, nugget coverage from 0.282 to 0.291, and TFC1 from 0.262 to 0.349, while retrieval precision decreases from 0.933 to 0.922. This pattern suggests that MiniLM improves answer-content matching and query-term evidence, but can select from a slightly less precise cited-document set.

Within the topic-shift family, MiniLM improves BERTScore from 0.156 to 0.163, retrieval precision from 0.929 to 0.940, and TFC1 from 0.215 to 0.373, but ROUGE-L decreases from 0.167 to 0.157 and nugget coverage is roughly similar, moving from 0.292 to 0.288. This suggests that MiniLM changes the focus and wording of selected evidence rather than uniformly improving every metric.

The rule-based family gives the clearest positive case for late sentence reranking. \texttt{rule-minilm} exceeds both \texttt{rrf-no-rerank} and \texttt{caes-rag-rrf} on BERTScore, retrieval precision, nugget coverage, average grade, and TFC1, while remaining close on ROUGE-L. This points to sentence-level allocation as a more effective intervention than redefining all chunks with embedding similarity in the current candidate-constrained setting.

\subsection{ROUGE-L Is Limited in Same-Retrieval Comparisons}
ROUGE-L remains useful as an end-to-end surface-form similarity metric, but it should not be interpreted as a clean measure of retrieval or evidence-selection quality. For example, \texttt{default} and \texttt{caes-rag-rrf} use effectively the same retrieval stack: rule-based chunking, multi-query expansion, PRF, RRF fusion, lightweight evidence reranking, and citation prior. Differences between them in ROUGE-L are therefore likely driven mainly by final answer wording rather than by genuinely different retrieval behavior.

This observation supports a broader methodological point: in same-retrieval comparisons, ROUGE-L variation can overstate generation-style differences. For analyzing core RAG capability, especially grounding and evidence quality, retrieval precision, nugget coverage, and average grade are more informative than ROUGE-L alone.

\section{Supplementary Diagnostic Results}
We now report the supplementary Claude Sonnet 4.5 judge evaluation as a diagnostic view rather than as the primary result. Table~\ref{tab:summary-by-model} shows the mean correctness, completeness, average score, and citation-set statistics for all ten compared methods across the 47 evaluation queries. Correctness and completeness are Claude Sonnet 4.5 judge rubric scores for factual consistency and information coverage, respectively; Avg. is their arithmetic mean. The three citation-set metrics compare reference-side cited documents $G$ with RAG-side cited documents $R$: Recall $=|G\cap R|/|G|$, Precision $=|G\cap R|/|R|$, and Jaccard $=|G\cap R|/|G\cup R|$ \cite{zheng2023llm_judge,liu2023geval,jaccard1901distribution}.

\begin{table}[t]
\centering
\scriptsize
\setlength{\tabcolsep}{3pt}
\caption{Supplementary diagnostic evaluation under the Claude Sonnet 4.5 judge. Citation-set metrics are reported as 0--1 ratios.}
\label{tab:summary-by-model}
\resizebox{\textwidth}{!}{%
\begin{tabular}{r l r r r r r r}
\toprule
No. & Approach & Corr. & Comp. & Avg. & Recall & Precision & Jaccard \\
\midrule
1 & \texttt{rrf-no-rerank} & 3.000 & 2.468 & 2.734 & 0.708 & 0.922 & 0.695 \\
2 & \texttt{rule-minilm} & 2.915 & 2.447 & 2.681 & 0.816 & 0.961 & 0.787 \\
3 & \texttt{single-query-bm25} & 2.979 & 2.383 & 2.681 & 0.761 & 0.940 & 0.742 \\
4 & \texttt{caes-rag-rrf} & 2.936 & 2.362 & 2.649 & 0.726 & 0.947 & 0.720 \\
5 & \texttt{default} & 2.915 & 2.362 & 2.639 & 0.747 & 0.936 & 0.738 \\
6 & \texttt{topic-shift-current} & 2.830 & 2.362 & 2.596 & 0.776 & 0.908 & 0.740 \\
7 & \texttt{semantic-current} & 2.787 & 2.404 & 2.595 & 0.763 & 0.938 & 0.735 \\
8 & \texttt{topic-shift-minilm} & 2.830 & 2.298 & 2.564 & 0.754 & 0.911 & 0.715 \\
9 & \texttt{semantic-minilm} & 2.787 & 2.340 & 2.563 & 0.797 & 0.904 & 0.724 \\
10 & \texttt{concat-baseline} & 1.574 & 1.319 & 1.446 & 0.132 & 0.238 & 0.124 \\
\bottomrule
\end{tabular}%
}
\end{table}

\subsection{Supplementary Evaluation Highlights Lexical Fusion}
Under the supplementary Claude Sonnet 4.5 judge protocol, \texttt{rrf\_no\_rerank} obtains the highest mean score, 2.734, followed by \texttt{rule\_minilm} and \texttt{single\_query\_bm25}, both at 2.681. This suggests that the local judge rewards broadly plausible answer quality from a diverse lexical-fusion evidence pool. However, because the primary evaluation favors \texttt{rule-minilm}, we interpret this local evaluation result as diagnostic rather than definitive.

The supplementary results still confirm an important lower-level point: retrieval and evidence selection are necessary. The \texttt{concat\_baseline} performs substantially worse than retrieval-based systems, with an average judge score of 1.446 and mean Jaccard coefficient of 0.124.

\subsection{Primary and Supplementary Evaluations Disagree Informatively}
The most important disagreement is that supplementary LLM-judge evaluation assigns the highest mean score to \texttt{rrf\_no\_rerank}, whereas primary metrics favor \texttt{rule-minilm}. This suggests that a single local judge can reward fluent or broadly plausible synthesis while underweighting improvements in cited-evidence precision and nugget coverage. The primary metrics separately expose lexical overlap, semantic similarity, citation precision, nugget coverage, average grade, and TFC1, so they provide a more detailed picture of where MiniLM sentence reranking helps.

\section{Discussion}
The primary LongEval-RAG results refine the main lesson of our study. Candidate-constrained provenance control is necessary, but not sufficient; the largest gains come from selecting answer-worthy evidence inside the fixed candidate set. Among our variants, the strongest primary-evaluation configuration is \texttt{rule-minilm}, which combines deterministic rule-based chunks with late MiniLM sentence reranking.

A likely explanation is division of labor. Rule chunks preserve local sentence context with predictable overlap, while query expansion, PRF, and RRF produce a diverse lexical candidate pool. Lightweight reranking and the citation prior keep evidence selection inside the official candidate set. MiniLM then operates late, where it can choose answer-worthy sentences without changing document segmentation. This may avoid error propagation from semantic or topic-shift chunking while still benefiting from neural matching.

The primary results do not imply that semantic or topic-shift chunking is intrinsically ineffective. The candidate pool contains only ten documents per query, so upstream chunk boundaries may matter less than final sentence allocation. Embedding-based chunking may also split or merge scientific text in locally coherent ways that are not optimized for the gold-answer nuggets. Scientific abstracts and full texts contain dense terminology, so lexical overlap and title/body signals may remain strong baselines. The mixed primary metrics show that semantic and topic-shift variants can improve some measures while harming others, which means they need tuning against the intended metric target.

The disagreement between primary and supplementary evaluations is also a useful result. Local LLM judging appears to reward broadly plausible answer quality, while the primary metrics separately measure gold-answer overlap, semantic similarity, cited-document precision, nugget coverage, and query-term constraints. This shows why RAG evaluation should combine multiple complementary measures rather than relying on a single judge score or a single lexical-overlap metric.

Future work should therefore focus on late-stage sentence allocation and metric-aware reranking. In particular, we should diagnose when MiniLM improves nugget coverage, when it changes only answer phrasing, and when it selects less precise citations. The fixed-10-document setting also calls for more explicit diagnostics separating cases where the right document is present but ranked too low from cases where the document is selected but not used well in the answer.

\section{Limitations}
Several limitations should be considered when interpreting these results. First, the current experiments use a 47-query evaluation slice, so small differences between closely ranked systems should be treated as suggestive rather than conclusive without additional queries, confidence intervals, or repeated runs. Second, primary metrics are more authoritative than our supplementary judge, but they still emphasize particular reference answers, nugget definitions, and metric implementations; alternative gold annotations could change fine-grained conclusions. Third, the supplementary reference-style answers are model-generated rather than written by human annotators, which means they may omit legitimate evidence, reflect model-specific synthesis preferences, or encode errors that affect downstream judging. Fourth, although the supplementary judge uses a different model family from the answer generator, LLM-as-judge evaluation can still be sensitive to prompt wording, answer style, verbosity, and subtle factual errors. Fifth, document-ID overlap is only a diagnostic proxy: high overlap does not guarantee that an answer uses evidence correctly, and low overlap does not necessarily imply an incorrect answer when several candidate documents contain similar information. Finally, the fixed ten-document candidate setting and the mixture of full-text and abstract-only inputs limit how directly these scores generalize to other RAG settings. We therefore interpret both the supplementary and primary numbers as protocol-dependent comparative evidence about pipeline behavior, not as an absolute estimate of final task performance.

\section{Conclusion}
The primary LongEval-RAG results refine the main lesson of our study. Candidate-constrained provenance control is necessary, but not sufficient; the largest gains come from selecting answer-worthy evidence inside the fixed candidate set. Among our variants, the strongest primary-evaluation configuration is \texttt{rule-minilm}, which combines deterministic rule-based chunks with late MiniLM sentence reranking. This method achieves the best BERTScore, retrieval precision, nugget coverage, and average grade among our submissions, suggesting that neural reranking is most useful when applied after stable lexical retrieval and passage construction.

At the same time, disagreements between supplementary LLM-judge evaluation and primary evaluation show that RAG systems should be assessed with multiple complementary measures rather than a single judge score. These results provide a clear next direction: keep the provenance-preserving candidate-constrained framework, retain the strongest rule-based and MiniLM sentence-reranking baselines, and focus future work on metric-aware late-stage evidence allocation.

\section*{Code Availability}
The implementation used for this LongEval-RAG system is available at \url{https://github.com/WiJoWill/longeval2026task4}.

\section*{Declaration of Generative AI and AI-assisted technologies}
Generative AI tools were used in experiment design and implementation, analysis, and manuscript
drafting and editing; all outputs were reviewed and revised by the authors, who take full responsibility
for the content.

\bibliography{references}

\end{document}